\documentclass[runningheads]{llncs}

\usepackage{eccv}

\usepackage{eccvabbrv}

\usepackage{graphicx}
\usepackage{booktabs}

\usepackage{adjustbox}
\usepackage{subcaption}
\usepackage{multirow}
\usepackage{float}

\usepackage[accsupp]{axessibility}  

\usepackage{hyperref}

\usepackage{orcidlink}

\begin{document}

\title{Motion-Guided Latent Diffusion for Temporally Consistent Real-world Video Super-resolution} 

\titlerunning{MGLD-VSR}

\author{
Xi Yang\inst{1,2} \and 
Chenhang He\inst{1} \and
Jianqi Ma\inst{1,2} \and
Lei Zhang\inst{1,2\footnotemark[1]}}

\authorrunning{X.~Yang et al.}

\institute{
The Hong Kong Polytechnic University, Hong Kong SAR, China  \and
OPPO Research Institute, Shenzhen, China \\
\email{\{xxxxi.yang, jianqi.ma\}@connect.polyu.hk}, 
\email{chenhang.he@polyu.edu.hk}, \\
\email{cslzhang@comp.polyu.edu.hk}
}

\maketitle

\renewcommand{\thefootnote}{\fnsymbol{footnote}}
\footnotetext[1]{Corresponding author. This work is supported by the Hong Kong RGC RIF grant (R5001-18) and the PolyU-OPPO Joint Innovation Lab.}

\begin{abstract}
  Real-world low-resolution (LR) videos have diverse and complex degradations, imposing great challenges on video super-resolution (VSR) algorithms to reproduce their high-resolution (HR) counterparts with high quality. Recently, the diffusion models have shown compelling performance in generating realistic details for image restoration tasks. However, the diffusion process has randomness, making it hard to control the contents of restored images. This issue becomes more serious when applying diffusion models to VSR tasks because temporal consistency is crucial to the perceptual quality of videos. In this paper, we propose an effective real-world VSR algorithm by leveraging the strength of pre-trained latent diffusion models. To ensure the content consistency among adjacent frames, we exploit the temporal dynamics in LR videos to guide the diffusion process by optimizing the latent sampling path with a motion-guided loss, ensuring that the generated HR video maintains a coherent and continuous visual flow. To further mitigate the discontinuity of generated details, we insert temporal module to the decoder and fine-tune it with an innovative sequence-oriented loss. The proposed motion-guided latent diffusion (MGLD) based VSR algorithm achieves significantly better perceptual quality than state-of-the-arts on real-world VSR benchmark datasets, validating the effectiveness of the proposed model design and training strategies. Codes and models are available at https://github.com/IanYeung/MGLD-VSR.
  \keywords{Video super-resolution \and Motion-guided latent diffusion}
\end{abstract}

\section{Introduction}

\begin{figure}[t]

\centering

\includegraphics[width=1.0\linewidth]{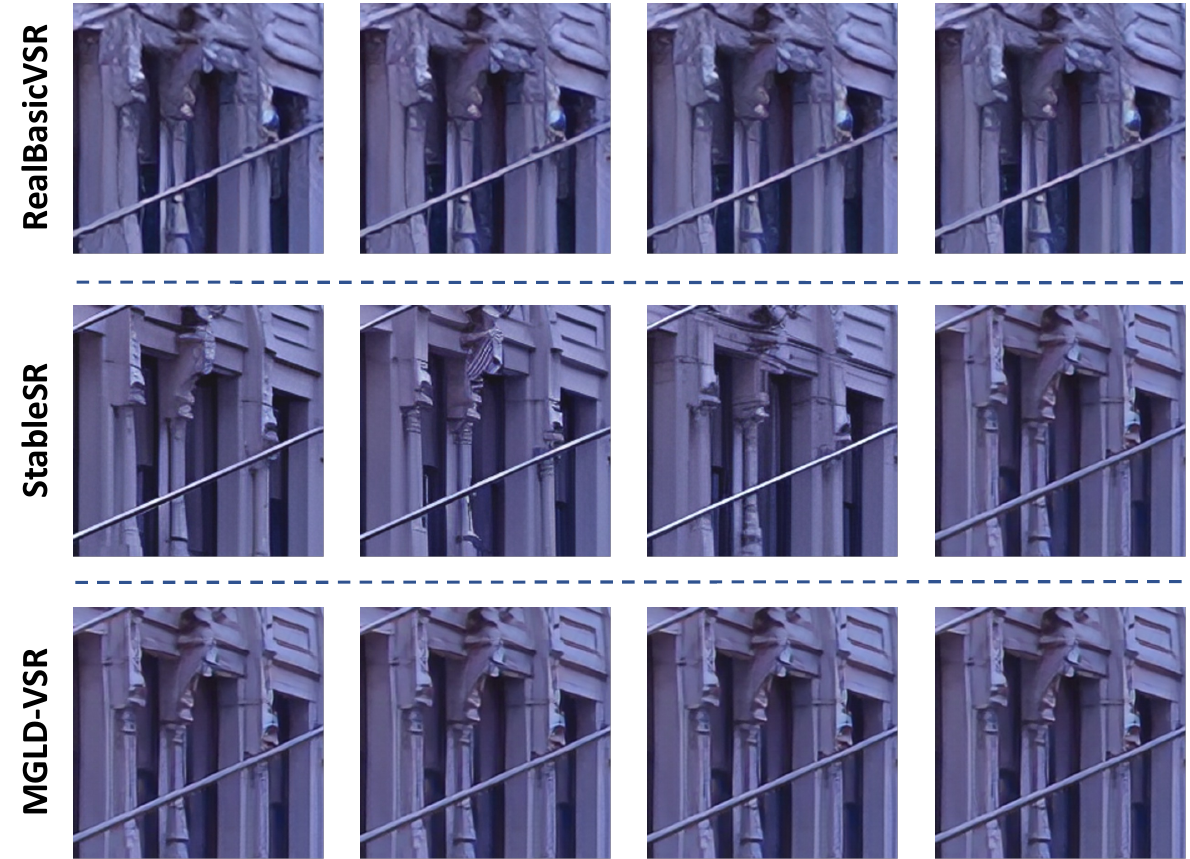}

\caption{The VSR results of 4 consecutive frames in the Sequence 026 from VideoLQ.  The top row shows the VSR results by RealBasicVSR \cite{chan2022investigating}, the middle row shows the naive VSR results by running StableSR \cite{wang2023exploiting} on each frame of the sequence, and the bottom row shows the VSR results produced by our proposed MGLD-VSR method. Our method generates realistic details while achieving good temporal consistency.}

\label{figure:motivation}
\end{figure}

Video super-resolution (VSR) \cite{bishop2003super} aims at reconstructing a high-resolution (HR) video from a given low-resolution (LR) video sequence. Thanks to the rapid development of deep learning techniques, significant progress has been made for VSR in the past decade. Representative VSR methods include sliding window based methods like EDVR \cite{wang2019edvr}, recurrent based methods like BasicVSR \cite{chan2021basicvsr} and more recent transformer based methods \cite{liang2022vrt}. However, most of the above works assume simple degradations (e.g., bicubic resizing, downsampling after Gaussian blur) between the LR and HR videos. As a result, the trained VSR models are hard to be generalized to real-world LR videos, whose degradations are unknown and much more complex.

Recently, real-world VSR \cite{yang2021real, chan2022investigating, xie2023mitigating} has gained attention from researchers for its high potential value in practical applications such as camera-phone video enhancement and online video streaming. Different from traditional VSR tasks, which focus on super-resolving clean LR videos with simple synthetic degradation (e.g., bicubic downsampling), real-world VSR aims to enhance videos with complex and unknown degradations that often appear in real-world low-quality videos. The main challenge lies in how to effectively reproduce video details while suppressing the visual artifacts caused by unknown degradations. Existing real-world VSR algorithms mostly attempt to trade-off between details and artifacts using techniques such as Laplacian pyramid generative adversarial loss \cite{yang2021real}, input pre-cleaning \cite{chan2022investigating} and hidden state attention \cite{xie2023mitigating}, etc. Despite these efforts and progresses, their performance on real-world testing sets remains limited due to the insufficient training data and the limited model capacity. 

Since the seminal work of denoising diffusion probabilistic model (DDPM) \cite{ho2020denoising}, diffusion-based generative models have achieved great success in image generation. In particular, the latent diffusion models (LDMs) \cite{rombach2021highresolution} have demonstrated compelling results not only in text-to-image generation, but also in downstream tasks such as image editing \cite{kawar2023imagic}, inpainting \cite{lugmayr2022repaint}, colorization \cite{carrillo2023diffusart}, etc. Recently, researchers have also attempted to employ the powerful generative priors of LDMs for real-world image restoration tasks \cite{wang2023exploiting} and shown promising results. It is thus interesting to investigate whether the latent diffusion priors can be used to improve real-world VSR results. However, this is a non-trivial task because of the intrinsic randomness in the diffusion process. This randomness in turn gives rise to the temporal inconsistencies among video frames in the latent space. Such inconsistencies can be further amplified by the decoder of the LDM. As shown in the middle row of Figure \ref{figure:motivation}, directly employing the LDM based image super-resolution algorithm \cite{wang2023exploiting} for VSR will lead to inconsistent details across consecutive frames, disturbing the visual perception of the reproduced video.

To address the above-mentioned challenges, we propose a Motion-Guided Latent Diffusion (MGLD) model for real-world VSR in this paper, aiming to produce high-quality HR video sequence with good temporal consistency. We propose to incorporate the motion dynamics of input LR video into the the generation of SR outputs. This can be done through a guided diffusion process where we can generate conditional samples from a guidance model based on its score function \cite{song2021scorebased, Ranzato2021diffusion}. In particular, we calculate the optical flows among adjacent LR frames, and use the calculated optical flows to warp the latent features of each frame to align the features of neighboring frames. The $l_1$ norm of the warping error is employed as the motion-guided loss, and its gradients can be incorporated in the sampling process to update the latent features.

To overcome the inconsistent details generated by the original decoder of LDM, we insert temporal modules to the decoder and fine-tune it with the ground-truth HR video sequences. We also introduce an innovative sequence-oriented loss to guide the decoder to improve the continuity of the details. As depicted in the bottom row of Figure \ref{figure:motivation}, our proposed MGLD-VSR approach could mitigate the discontinuity of generated details while maintaining natural and visually appealing texture recovery.

The contributions of this paper are summarized as follows. First, we propose a diffusion sampling process based on motion-guided loss, allowing the temporal dynamics of the input frame to be used in generating temporally consistent latent features. Second, we propose a temporal-aware sequence decoder, along with two sequence-oriented losses, to further enhance the continuity of generated videos. Our proposed MGLD-VSR model achieves highly competitive real-world VSR results, exhibiting perceptually much more realistic details with fewer flickering artifacts than existing state-of-the-arts.

\section{Related Work}
\label{sec:relatedwork}

\textbf{Video Super-Resolution.} 
The recent advances in VSR research largely attribute to the rapid development of deep-learning technologies \cite{caballero2017real, jo2018deep, wang2019edvr, chan2021basicvsr, liang2022vrt}. Based on the paradigms, existing deep-learning based VSR algorithms could be roughly classified into two categories. 
The first category performs VSR in a multiple-input single-output (MISO) fashion. Typical algorithms include DUF \cite{jo2018deep},  EDVR \cite{wang2019edvr}, and MuCAN \cite{li2020mucan}, etc. The MISO algorithms are usually sliding-window based, which aggregate information from neighboring frames to generate the super-resolved result of the center frame. 
The other category of methods adopt a multiple-input multiple-output (MIMO) design, such as BasicVSR \cite{chan2021basicvsr}, BasicVSR++ \cite{chan2022basicvsr++}, VRT \cite{liang2022vrt} and RVRT \cite{liang2022recurrent}. These methods take multiple LR frames as input and generate their corresponding SR results in one go. Compared with MISO, the amortized running cost of MIMO is usually much lower as it avoids a series of redundant operations. In addition, benefiting from the lower cost, the MIMO methods are able to aggregate information from more frames to improve the performance. We adopt the MIMO design in this work.

\textbf{Real-world VSR.}
Most existing VSR methods are trained on synthetic data with simple degradations (e.g., bicubic resizing, downsampling after Gaussian blur). Due to the synthetic-to-real gap, models trained with such data cannot perform well on real-world videos. To tackle this problem, Yang \textit{et al.} \cite{yang2021real} collected LR-HR video pairs with iPhone cameras to better model real-world degradations. While being effective in some cases, it is relatively labor-intensive and may not generalize well to videos collected by other devices. Later works usually adopt an easier solution: synthesizing more realistic training data by mixing more types of degradations, such as blur, downsampling, noise and video compression \cite{chan2022investigating, xie2023mitigating}. With the improved synthetic training data, the trained models could achieve better super-resolution performance on real-world videos. However, there is still much distribution gap between the synthetic LR data and real-world LR videos. Instead of working on synthesizing more realistic training data, in this work we investigate how to leverage the strong generative priors of pre-trained latent diffusion models to improve the real-world VSR performance.

\textbf{Generative Prior for Image Restoration.}
The highly ill-posed nature of real-world restoration problem brings forward a great challenge on real-world image restoration algorithms. To boost the performance, a series of methods have introduced generative priors into the image restoration process. Early methods usually adopt pre-trained generative adversarial networks (GANs) as the generative prior \cite{menon2020pulse, gu2020image, wang2021towards, Yang2021GPEN, chan2021glean}. Although effective, the priors adopted in these methods are usually tailored for specific tasks (e.g., face image restoration), limiting their capability in handling more general natural image restoration problems. Considering the higher generality of diffusion models than GANs in natural image generation, recently researchers have started to investigate how to leverage generative diffusion priors for image restoration, and demonstrated interesting results \cite{fei2023generative, wang2023exploiting}. In this work, we investigate the possibility of applying latent diffusion priors to real-world VSR, and develop effective techniques to tackle the flickering issue and improve the temporal consistency of generated videos.

\textbf{Diffusion for Video Synthesis.} 
Recently, diffusion based models have demonstrated its superiority in the video synthesis \cite{singer2022make, blattmann2023align, ge2023preserve, blattmann2023stable}. As one of the pioneer works, Make-a-Video \cite{singer2022make} leverages a pretrained text-to-image (T2I) model and build a text-to-video (T2V) model by inserting temporal modules into the T2I model and designing a cascaded architecture. VideoLDM \cite{blattmann2023align} adopts a latent diffusion paradigm and train the T2V model on large-scale paired text-video datasets, greatly improving the generation results. Stable Video Diffusion \cite{blattmann2023stable} further improve the results by conducting training on more carefully constructed datasets and introducing more advanced training paradigm. In this work, we leverage the diffusion prior and build a powerful real-world VSR model, which aims to tackle the ill-poseness of real-world VSR problem with diffusion prior and obtain temporally consistent VSR results with rich details.

\begin{figure}[t]
\begin{center}

\includegraphics[width=1.0\linewidth]{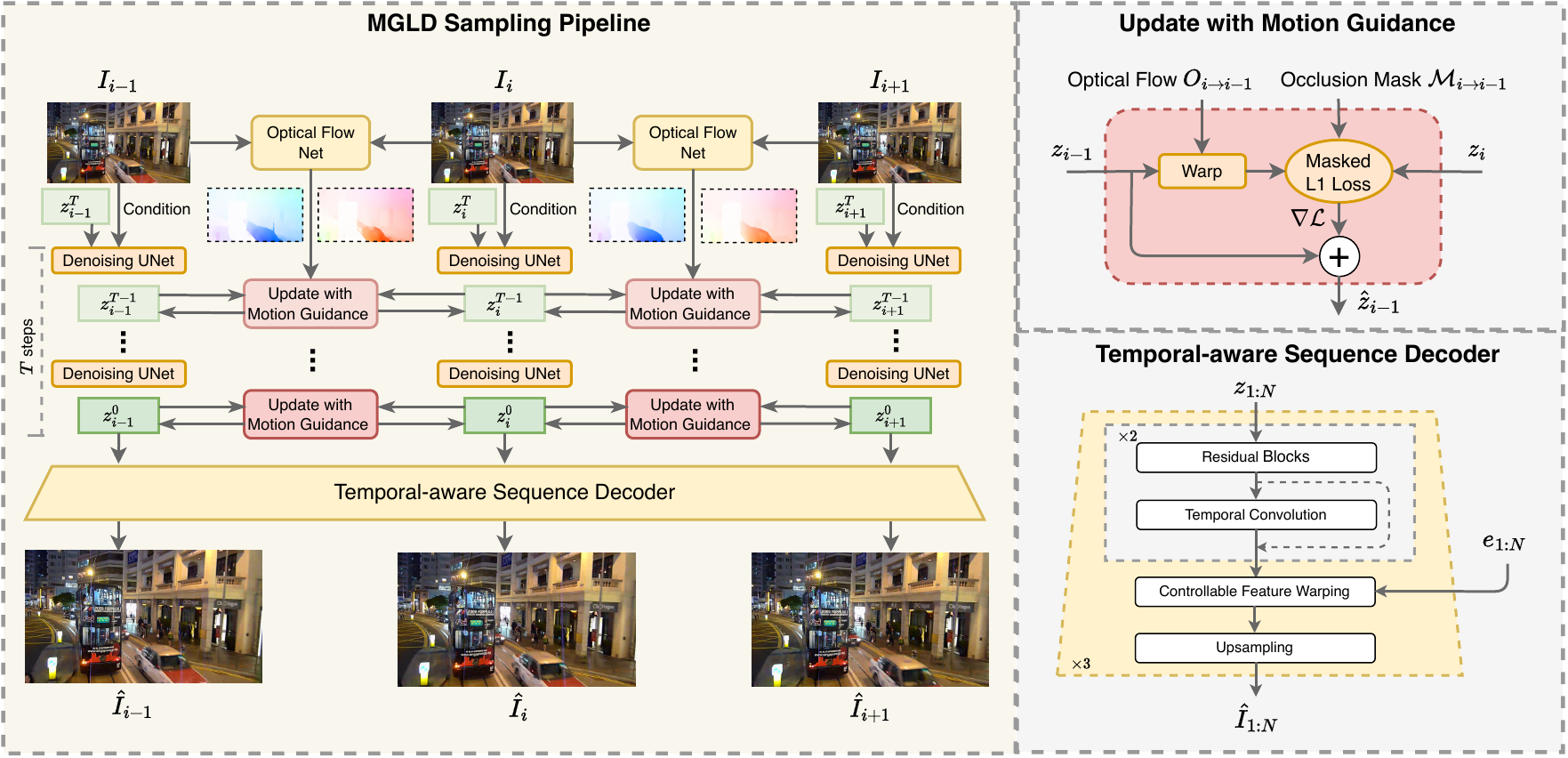}
\end{center}

\caption{Overview of the proposed MGLD-VSR framework for real-world VSR. We first estimate the forward and backward optical flow with a pre-trained optical flow estimation network and then employ a motion-guided loss to update the latent diffusion sampling process. The motion-guided loss is computed by summing the forward and backward masked warping error of the latent sequence at each time-step. After a number of $T$ sampling steps, we obtain the generated latent sequence and feed it into the temporal-aware sequence decoder to reconstruct the VSR sequence.}
\label{fig:overview}

\end{figure}

\section{Methodology}

Given an LR video sequence of $N$ frames $I = \{I_1, I_2, \ldots, I_N\}$, our goal is to reproduce an HR video sequence $\hat{I} = \{\hat{I}_1, \hat{I}_2, \ldots, \hat{I}_N\}$ with enhanced visual quality while maintaining the content consistency across adjacent frames. 
Inspired by the recently developed diffusion-based image restoration methods \cite{fei2023generative, wang2023exploiting}, we propose to harness the generative priors of the pre-trained SD model \cite{rombach2021highresolution} to improve the real-world VSR performance. 
First, we propose a motion-guided diffusion sampling process, where the temporal dynamics from the LR sequence are introduced into the diffusion sampling process to enhance the consistency of latent features across frames. 
Then, we further improve the quality of generated details by designing a temporal-aware sequence decoder and fine-tuning it with a video-oriented loss. 
The overall framework of our algorithm, namely motion-guided latent diffusion (MGLD), is illustrated in Figure \ref{fig:overview}. 

\subsection{Latent Diffusion Model}
Diffusion models \cite{NEURIPS2020_4c5bcfec} are a class of generative models which transform Gaussian noise into data samples through an iterative reverse Markovian process. This process can be described by:
$q(x_{t} | x_{t-1}) := \mathcal{N}(x_{t}, (1 - \beta_{t})x_{t-1}, \beta_{t}I),$
where the Gaussian noise is iteratively applied to the sample
$x_{t-1}$ at time-step $t-1$ to produce $x_t$, guided by a variance scheduler $\{\beta_{t}\}_{t = 1}^T$ along the diffusion process. The key objective of diffusion models is to train a denoising U-Net $\epsilon_{\theta}$ parameterized by $\theta$ and conditioned on the time-step $t$ and an optional signal $y$. Given the noise target $\epsilon_t$, the diffusion loss can be defined as:
$L(\theta) = \mathbb{E}_{t \sim U(1, T), \epsilon_t \sim \mathcal{N}(0, I)} \left\| \epsilon_t - \epsilon_{\theta}(\mathbf{x}_t; t, y) \right\|_2^2.$

Latent Diffusion Models (LDMs) \cite{rombach2021highresolution} have made a remarkable impact in the field of image generation. These models employ Variational Autoencoders (VAE) to map the images into a latent space, enabling them to be trained on large-scale text-image datasets. This equips LDMs with powerful prior knowledge about natural images,  which can be leveraged by downstream tasks such as image restoration and super-resolution \cite{wang2023exploiting} to produce images with clear details and high-quality content. However, in the case of VSR, the stochastic nature of diffusion process often leads to the temporal inconsistencies in the restored video sequence. To solve this issue, we propose a MGLD model that achieves high-quality VSR by utilizing guidance during the sampling process and incorporating temporal awareness in the decoder.

\subsection{Motion-guided Diffusion Sampling}\label{sec:MGS}
From a score-based perspective \cite{song2021scorebased}, the sampling process of a diffusion model involves the application of Langevin Dynamics to both gradient of the data distribution $\log p(\mathbf{z})$ and a guidance model $g = G(z)$, which can be a network \cite{NEURIPS2021_49ad23d1, pmlr-v162-nichol22a} or an energy model \cite{chen2023trainingfree}:
\begin{equation}
    \nabla_{\mathbf{z}} \log p(\mathbf{z}, g) = \nabla_{\mathbf{z}} \log p(\mathbf{z}) + \nabla_{\mathbf{z}} \log p(g|\mathbf{z}).
    \label{eq:g}
\end{equation}

In this work, we introduce a technique that utilizes the temporal dynamics in low-resolution input videos to create latent features that maintain temporal consistency, thereby enhancing VSR results. Specifically, we incorporate an innovative motion-guided module within the sampling process to measure the warping errors of latent features across frames. Given the LR frames, we first compute their optical flow with $\mathcal{F}$ \cite{teed2020raft} and downsample the flow map to the dimension of latent features. Given the forward flow $O_{f}$ and backward flow $O_{b}$, we warp the latent features to their neighbouring frames and compute the accumulated warping error along both directions:
\begin{align}
\begin{split}
    E(z)^t & = \textstyle\sum_{i=1}^{N-1}\Vert  (\operatorname{Warp}(z^t_{i}, O_{b,i})-z^t_{i+1})) \Vert_1 \\
      & + \textstyle\sum_{i=2}^{N}\Vert  (\operatorname{Warp}(z^t_{i}, O_{f,i-1})-z^t_{i-1})) \Vert_1.
\end{split}
\end{align}

We empirically find that the presence of occlusion will pose adverse impact on flow estimation, which in turn causes distraction in the sampling process and leads to artifacts in the final VSR results. To solve this issue, we further estimate the occlusion mask $\mathcal{M}$ in each frame \cite{meister2018unflow} and ignore the contributions in the occluded region. Given the variance $\sigma_t^2$ of noise scheduler at step $t$, the motion-guided sampling process at sampling step $t$ can be written as:
\begin{equation}
\hat{z}^{t} \leftarrow \text{DDPM}(z^{t+1}, t) - \eta {\sigma}_{t}^2 {\nabla}_{z} (\mathcal{M} \circ  E(z)^t), 
\end{equation}
where the DDPM \cite{NEURIPS2020_4c5bcfec} scheduler is applied for diffusion reverse (denoising) process and $\eta=10$ is a weighting factor to control the guidance strength. After $T$ iterations, the final denoised latent sequence $z^{0}$ is fed into the VAE decoder to obtain the final video sequence.

\subsection{Temporal-aware Decoder Fine-tuning}\label{sec:TSD}
With the proposed motion-guided diffusion sampling, the VSR model could better exploit the temporal dynamics to generate temporally more consistent HR videos. However, since the guidance is performed in the low-resolution latent space, which is 8 times smaller in both height and width than the output image space, the video reconstruction through a VAE decoder may still lead to temporal discontinuities in the generated details of VSR output. Thus, we introduce a temporal-aware sequence decoder and fine-tune it with ground-truth video sequences to improve the smoothness and accuracy of output videos.

As shown in Figure \ref{fig:overview}, we build our sequence decoder on top of the pre-trained VAE decoder by incorporating 1D convolutions along the temporal dimension. This design facilitates interleaved spatial-temporal interactions and enables our module to restore details with high continuity while incurring minimal computational cost. We further incorporate the information $e$ from the VAE encoder through a Controllable Feature Warping (CFW) module \cite{zhou2022codeformer, wang2023exploiting} for better restoration and generation results. To leverage the restorative power of the pretrained VAE, we freeze their original spatial blocks and only update the parameters in the temporal convolution and CFW module. 

We first employ the $\ell_1$ loss and perceptual loss \cite{johnson2016perceptual, zhang2018unreasonable} as the reconstruction loss $L_{\mathrm{recon}}$ for each frame, then utilize the GAN-loss and frame difference loss that tailored for video sequences. The frame difference loss $L_{\mathrm{diff}}$ calculates the difference between consecutive predicted frames $\hat{I}$ and ground truth frames $I^{gt}$, which is defined as:
\begin{equation}
    L_{\mathrm{diff}} = \textstyle\sum \Vert (\hat{I}_{i+1} - \hat{I}_{i}) - (I_{i+1}^{gt} - I_{i}^{gt}) \Vert_{1}.
\end{equation}
It is noteworthy that diffusion models typically generate strong structures. To make the super-resolved video look more natural, we need to apply consistency constraints to those structured regions. We hence introduce a structure weighted consistency loss $L_{\mathrm{swc}}$:
\begin{align}
\begin{split}
    L_{\mathrm{swc}} & = \textstyle\sum \Vert \mathcal{M} \circ (\mathcal{W}_{i+1} ( \operatorname{Warp}(\hat{I}_{i}, O_{b,i}^{gt})- \hat{I}_{i+1})) \Vert_1 \\
    & + \textstyle\sum \Vert \mathcal{M} \circ (\mathcal{W}_{i-1} ( \operatorname{Warp}(\hat{I}_{i}, O_{f,i}^{gt})- \hat{I}_{i-1})) \Vert_1.
\end{split}
\end{align}
where $O_{f}^{gt}$, $O_{b}^{gt}$ denote the forward and backward flow computed on the ground-truth sequence. This loss helps the network to generate consistent details by modulating the loss through a weighted map $\mathcal{W}=(1+wS)$  computed on the ground-truth frames, where $w=3$ is a weighting factor and $S$ is a structure (edge) map using the Sobel operator. Occlusion masking is also adopted. Given the GAN-based loss $L_{\mathrm{GAN}}$ and reconstruction loss $L_{\mathrm{recon}}$, the total loss for decoder fine-tuning is defined as: 
\begin{equation}
L_{\mathrm{video}} = L_{\mathrm{recon}} + \alpha L_{\mathrm{diff}} + \beta L_{\mathrm{swc}} + \gamma L_{\mathrm{GAN}},
\end{equation}
where $\alpha$, $\beta$ and $\gamma$ are empirically set to 0.5, 0.5 and 0.025.

\section{Experiments}
\subsection{Experimental Settings}

\noindent
\textbf{Implementation Details.}
The training of MGLD-VSR consists of two main stages. In the first stage, we fine-tune the denoising U-Net for latent space diffusion. The weight of the denoising U-Net is initialized from Stable Diffusion V2.1. We insert 1D temporal convolution into the U-Net to help temporal dynamic modeling. We then fix the weight in the denoising U-Net of SD model and train the conditioning and temporal modeling modules. For conditioning, we adopt a small time-aware encoder to encode LR condition and inject it into the denoising U-Net with spatial feature transform operation as in \cite{wang2023exploiting}. The training batch size, sequence length and latent image size are set to 24, 6 and $64 \times 64$, respectively. 
In the second stage, we first generate the clean latent sequence with the proposed motion-guided diffusion sampling process, then fine-tune the temporal-aware sequence decoder with the LR sequence, the generated latent sequence and the HR sequence. We fix the VAE decoder and insert temporal modeling layers and CFW modules for training. The training batch size, sequence length and image size are set to 4, 5 and $512 \times 512$, respectively. 

Our models are trained on 4 NVIDIA A100 GPUs with the PyTorch \cite{paszke2019pytorch} framework. Adam \cite{kingma2014adam} is chosen as the optimizer. During  inference, due to memory constrains, we divide the LR video into multiple sequences to run. To handle output with arbitrary resolution, we apply a progressive patch aggregation sampling algorithm \cite{wang2023exploiting}. For each sequence, we run the sampling for 50 steps.

\textbf{Training and Testing Datasets.}
As in previous works \cite{wang2019edvr,chan2022investigating, xie2023mitigating}, we merge the training set and validation set of REDS \cite{nah2019ntire} in training and leave 4 sequences for testing (REDS4). We follow the degradation pipeline of RealBasicVSR \cite{chan2022investigating} to synthesize the training sequence pairs, which involves blur, noise, downsampling and compression degradations.
We evaluate our method on both synthetic and real-world datasets. To synthesize LR-HR testing sequence pairs, we apply the RealBasicVSR degradation pipeline on several widely used VSR datasets, including REDS4 \cite{nah2019ntire}, UDM10 \cite{yi2019progressive} and SPMCS \cite{tao2017detail}. 
REDS4 contains 4 video sequences, each having 100 frames. UDM10 consists of 10 sequences, each having 32 frames. SPMCS has 30 sequences, each having 31 frames. 
For real-world dataset, we adopt VideoLQ \cite{chan2022investigating} for testing, which contains 50 real-world sequences with complex degradations.

\textbf{Evaluation Metrics.}
On synthetic datasets, we could adopt full-reference metrics to evaluate different real-world VSR methods. The perceptual quality metrics LPIPS \cite{zhang2018unreasonable} and DISTS \cite{ding2020image} are selected. Fidelity-oriented metrics PSNR and SSIM are also calculated for reference. Apart from image-based metrics, the video quality assessment metric VMAF \cite{li2018vmaf} is also employed.

On real-world datasets, as ground truths are not available, we compute the no-reference image quality metrics NIQE \cite{mittal2012making} and BRISQUE \cite{mittal2012no} for evaluation. In addition, we also include a deep learning based image-based metric MUSIQ \cite{ke2021musiq} and a state-of-the-art video quality assessment metric DOVER \cite{wu2023exploring}. 

\subsection{Experimental Results}
To demonstrate the effectiveness of the our MGLD-VSR algorithm, we compare it with seven state-of-the-art methods, including three real-world image super-resolution models (BSRGAN \cite{zhang2021designing}, RealESRGAN \cite{wang2021realesrgan}, StableSR \cite{wang2023exploiting}), one classic VSR model (RVRT \cite{liang2022recurrent}), and three blind or real-world VSR models (DBVSR \cite{pan2021deep}, RealVSR \cite{yang2021real}, RealBasicVSR \cite{chan2022investigating}. 

\textbf{Quantitative Comparison.}
We first show the quantitative comparison on the sythetic datasets and real-world video benchmarks in Table \ref{table:quantitative-results}. As shown from Table \ref{table:quantitative-results}, our method achieves the best results in term of full-reference perceptual based metrics LPIPS and DISTS on all synthetic test datasets, indicating that the proposed method is able to reconstruct high-quality plausible details from sequences degraded with complex degradations. Although methods like DBVSR may achieve better performance in terms of PSNR or SSIM on some datasets, they tend to generate blurry results, as demonstrated in the LPIPS and DISTS metrics. In terms of VMAF, our model achieves the best or second best results. As for real-world VSR dataset VideoLQ, our methods achieves the best performance in terms of NIQE, BRISQUE, and DOVER and the second best results in terms of MUSIQ, which reflects its strong capability to enhance the real-world videos and generate realistic details and textures.

\begin{table*}[!ht]
\begin{center}

\caption{
Quantitative comparison of different VSR models evaluated on popular VSR benchmarking datasets. \textbf{Bold} and \underline{underline} represent the best and second best score, respectively. For synthetic datasets (REDS4, UDM10, SPMCS), we adopt full-reference metrics to evaluate different methods. For real-world video datasets (VideoLQ), we adopt no-reference metrics to compare different methods. 
} 
\label{table:quantitative-results}

\begin{adjustbox}{width=1.0\linewidth}
\begin{tabular}{c|c|c|c|c|c|c|c|c|c|c}
\hline
Datasets & Metrics & Bicubic & RVRT & BSRGAN & RealESRGAN & StableSR & DBVSR & RealVSR & RealBasicVSR & Ours \\
\hline\hline
\multirow{5}{*}{REDS4} & PSNR $\uparrow$ & 22.13 & 22.29 & 22.02 & 21.34 & 22.04 & 22.29 & 21.78 & \underline{22.35} & \textbf{22.46} \\

& SSIM $\uparrow$ & 0.5730 & 0.5759 & 0.5702 & 0.5647 & 0.5683 & \underline{0.5760} & 0.5663 & \textbf{0.5846} & 0.5723 \\

& LPIPS $\downarrow$ & 0.6277 & 0.6221 & 0.4802 & 0.4509 & 0.4553 & 0.6221 & 0.6267 & \underline{0.4190} & \textbf{0.3776}  \\
   
& DISTS $\downarrow$ & 0.3667 & 0.3554 & 0.2118 & 0.1750 & 0.1845 & 0.3556 & 0.3620 & \underline{0.1581} & \textbf{0.1151}  \\

& VMAF $\uparrow$ & 2.1255 & 4.8005 & 26.2413 & 33.9955 & 26.6953 & 2.6075 & 2.5903 & \textbf{34.9884} & \underline{34.3214} \\

\hline\hline

\multirow{5}{*}{UDM10} & PSNR $\uparrow$ & 25.78 & \textbf{26.18} & 25.29 & 24.94 & 25.58 & \underline{26.14} & 25.09 & 25.56 & 25.99 \\

& SSIM $\uparrow$ & 0.7559 & \underline{0.7629} & 0.7323 & 0.7439 & 0.7592 & \textbf{0.7635} & 0.7524 & 0.7448 & 0.7548 \\

& LPIPS $\downarrow$ & 0.4441 & 0.4413 & 0.4217 & 0.3884 & \underline{0.3684} & 0.4398 & 0.4330 & 0.3897 & \textbf{0.3491}  \\
   
& DISTS $\downarrow$ & 0.2730 & 0.2575 & 0.1926 & 0.1651 & \underline{0.1497} & 0.2585 & 0.2608 & 0.1637 & \textbf{0.1369}  \\

& VMAF $\uparrow$ & 5.4647 & 10.1724 & 31.0874 & 36.9158 & 33.8264 & 7.2048 & 6.0079 & \underline{38.2638} & \textbf{39.3892} \\

\hline\hline

\multirow{5}{*}{SPMCS} & PSNR $\uparrow$ & 22.28 & \underline{22.51} & 22.12 & 21.66 & 22.26 & 22.50 & 22.03 & 22.41 & \textbf{22.66} \\

& SSIM $\uparrow$ & 0.5691 & 0.5753 & 0.5730 & 0.5694 & \underline{0.5844} & 0.5764 & 0.5738 & 0.5736 & \textbf{0.5960} \\

& LPIPS $\downarrow$ & 0.5572 & 0.5471 & 0.4509 & 0.4435 & \underline{0.4321} & 0.5458 & 0.5487 & 0.4420 & \textbf{0.3990}  \\
   
& DISTS $\downarrow$ & 0.3593 & 0.3415 & 0.2307 & 0.2150 & 0.2351 & 0.3435 & 0.3479 & \underline{0.2130} & \textbf{0.1934}  \\

& VMAF $\uparrow$ & 0.9656 & 5.3533 & 22.4828 & \underline{27.6935} & 21.8060 & 3.7576 & 2.7024 & 27.4519 & \textbf{28.7032} \\

\hline\hline

\multirow{4}{*}{VideoLQ} & NIQE $\downarrow$ & 9.1319 & 6.5888 & 4.2104 & 4.2036 & 4.4442 & 6.7504 & 6.3331 & \underline{3.6915} & \textbf{3.5346}  \\
   
& BRISQUE $\downarrow$ & 72.7405 & 62.1768 & 25.2400 & 29.8428 & 26.3092 & 61.1897 & 57.2765 & \underline{24.5484} & \textbf{21.9839}  \\

& MUSIQ $\uparrow$ & 21.6789 & 28.4204 & 52.2213 & 49.8380 & 47.8045 & 29.0039 & 24.2189 & \textbf{55.9691} & \underline{52.7812}  \\

& DOVER $\uparrow$ & 0.3677 & 0.4907 & 0.7184 & 0.7309 & 0.7108 & 0.4998 & 0.5202 & \underline{0.7401} & \textbf{0.7481}  \\

\hline
\end{tabular}
\end{adjustbox}

\end{center}
\end{table*}

\begin{figure*}[t]

\centering
\includegraphics[width=1.0\linewidth]{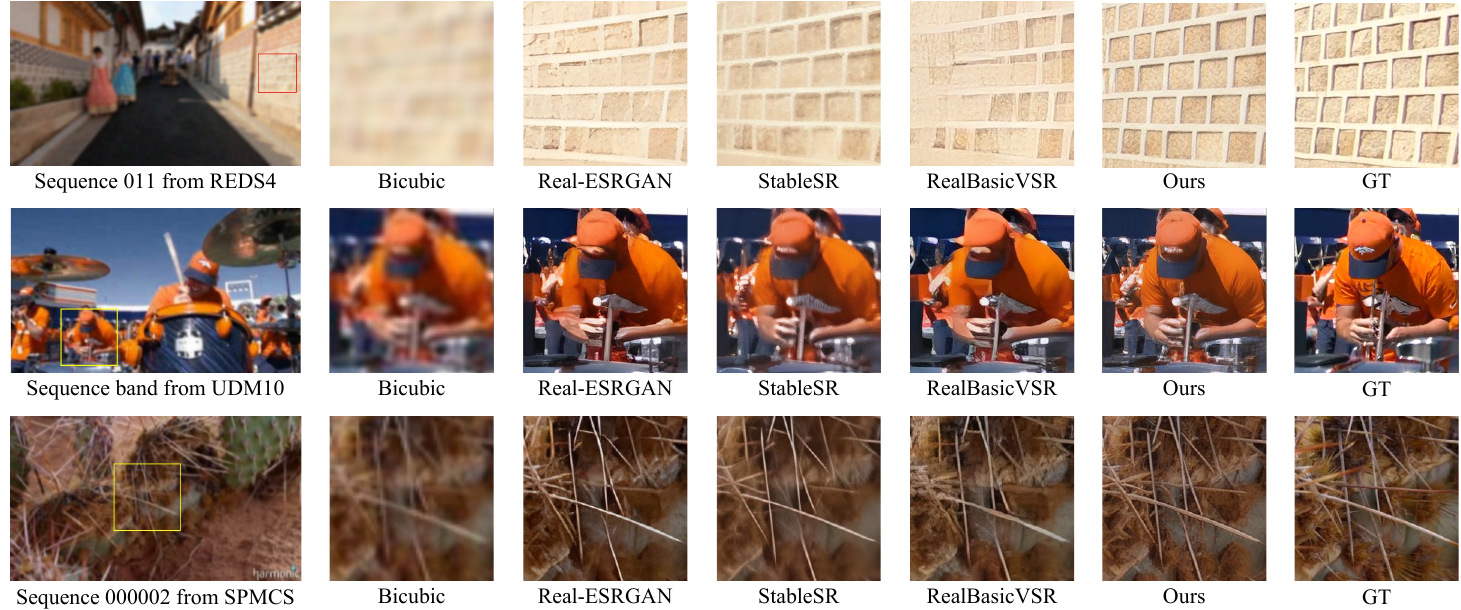}

\caption{Qualitative comparison on synthetic datasets (REDS4, UDM10, SPMCS) for $\times 4$ VSR. (Zoom-in for best view.)}

\label{figure:visual-comparison-syn}
\end{figure*}

\begin{figure*}[t]

\centering
\includegraphics[width=1.0\linewidth]{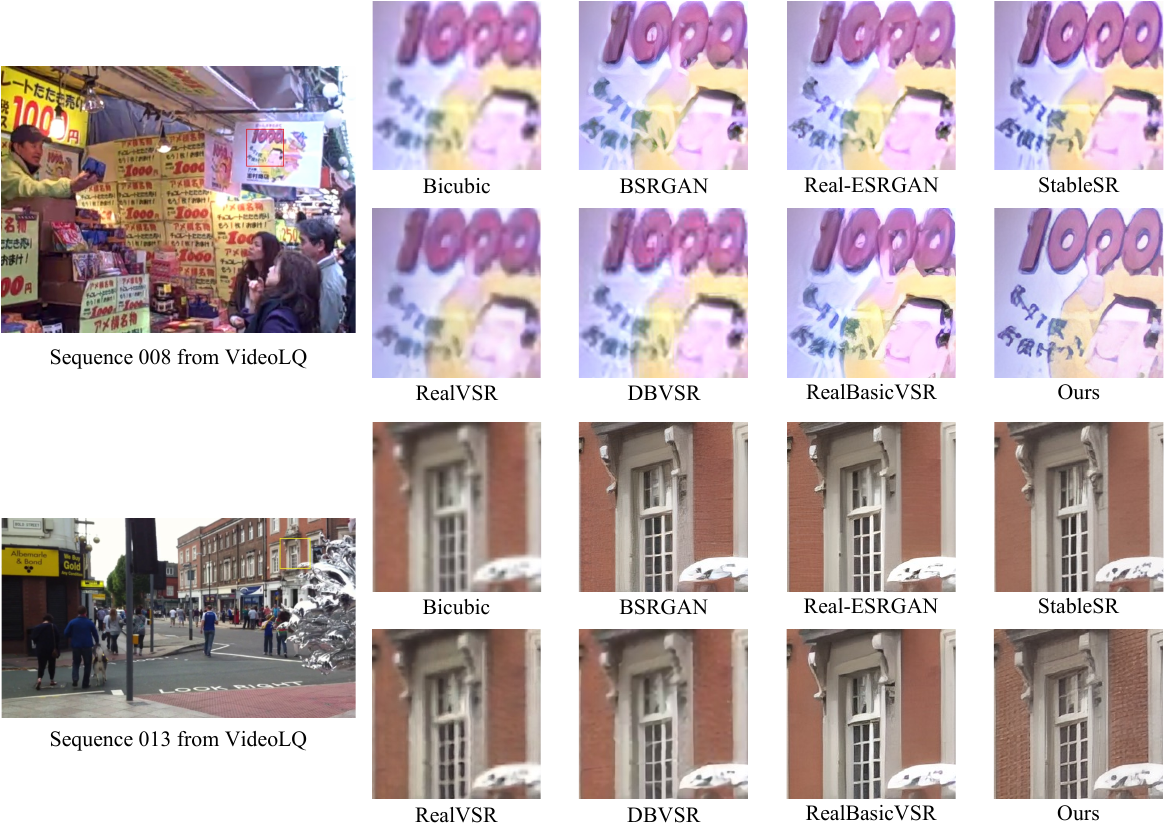}

\caption{Qualitative comparison on the real-world video dataset (VideoLQ) for $\times 4$ video SR. (Zoom-in for best view.)}

\label{figure:visual-comparison-real}
\end{figure*}

\textbf{Qualitative Comparison.}
To demonstrate the effectiveness of MGLD-VSR, we visualize the VSR results on synthetic datasets in Figure \ref{figure:visual-comparison-syn} and the results on real-world VideoLQ in Figure \ref{figure:visual-comparison-real}. One can see that MGLD-VSR can remove complex spatial-variant degradations and generate realistic details, outperforming other state-of-the-art real-world VSR algorithms. On synthetic datasets, as shown in Figure \ref{figure:visual-comparison-syn}, MGLD-VSR can well reconstruct the structure while generating plausible details under complex degradations. On real-world videos, as shown by the example of sequence 008 in Figure \ref{figure:visual-comparison-real}, MGLD-VSR reconstructs clearer characters and patterns than competing methods. For the example of sequence 013, MGLD-VSR not only reconstructs the structure of the windows, and at the same time generates plausible exterior structure of the brick wall. More visual comparisons can be found in the \textbf{supplementary materials}.

\begin{figure*}[!htbp]
\begin{center}
\includegraphics[width=1.0\linewidth]{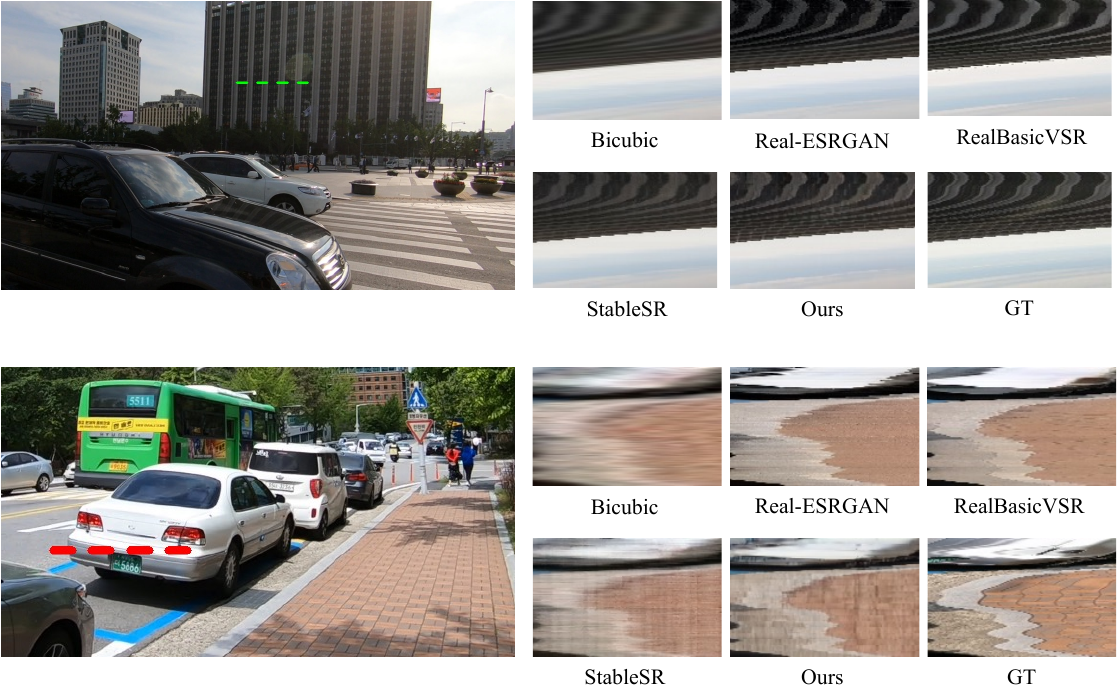}
\end{center}

\caption{Temporal profiles of competing real-world VSR methods.}
\label{figure:temporal-profile}

\end{figure*}

\textbf{Temporal Consistency.}
Compared with single image super-resolution, one major aspect of VSR is the temporal consistency of the generated videos. Following most previous works, in this paper, we adopt the average warping error (WE) \cite{lai2018learning, chu2020learning} of the sequence to quantitatively measure the temporal consistency:
$\operatorname{WE} = \tfrac{1}{N-1}\textstyle\sum_{i=1}^{N-1} \Vert \hat{I}_{i+1} - \mathrm{Warp}(\hat{I}_{i}, O_{b,i})\Vert_1,$
where $\hat{I}_{i}$ is the predicted frame and $O_{b,i}$ is the estimated backward optical flow \cite{xu2021high} from ground-truth video. On real-world video datasets without ground-truth, the optical flow is computed on the predicted frames.   

It is worth mentioning that, despite the wide use of WE for temporal consistency evaluation, it may not be able to faithfully reflect the real human perception on the video. For example, WE could be easily attacked by blurring the sequence, resulting much higher WE scores but lower video quality.
The WE scores of competing methods are given in Table \ref{table:quantitative-results-1}. One can see that our diffusion prior based method MGLD-VSR is not good at this metric because it tends to generate more details and textures, which are not favourable by WE. Nonetheless, the WE of MGLD-VSR is better than StableSR, which is also driven by the latent diffusion priors. To better showcase the effectiveness of our method, we visualize the temporal profiles of the VSR results produced by different algorithms in Figure \ref{figure:temporal-profile}. One can see that MGLD-VSR reconstructs more accurate details while maintaining good temporal consistency.

\begin{table}[!t]
\begin{center}
\caption{Temporal consistency comparison of competing VSR methods by using WE $\downarrow$. \textbf{Bold} and \underline{underline} represent the best and second best scores, respectively.} 
\label{table:quantitative-results-1}
\begin{adjustbox}{width=1.0\linewidth}
\begin{tabular}{c||c|c|c|c|c|c|c|c|c}
\hline
Datasets & Bicubic & RVRT & BSRGAN & RealESRGAN & StableSR & DBVSR & RealVSR & RealBasicVSR & Ours \\
\hline\hline
REDS4 & 8.57 & \textbf{6.50} & 10.25 & 11.91 & 11.81 & \underline{8.51} & 8.61 & 10.09 & 10.08  \\
UDM10 & 3.10 & 3.06 & 4.82 & 4.99 & 4.80 & \underline{3.04} & \textbf{2.75} & 4.33 & 4.55  \\
SPMCS & \underline{3.01} & 3.58 & 5.18 & 7.12 & 4.62 & 3.58 & 4.18 & \textbf{2.72} & 3.87 \\
VideoLQ & \textbf{5.48} & 6.38 & 7.72 & 7.64 & 9.33 & 5.98 & \underline{5.74} & 6.76 & 7.56  \\
\hline
\end{tabular}
\end{adjustbox}
\end{center}
\end{table}

\begin{figure}[!htbp]
\centering
\includegraphics[width=1.0\linewidth]{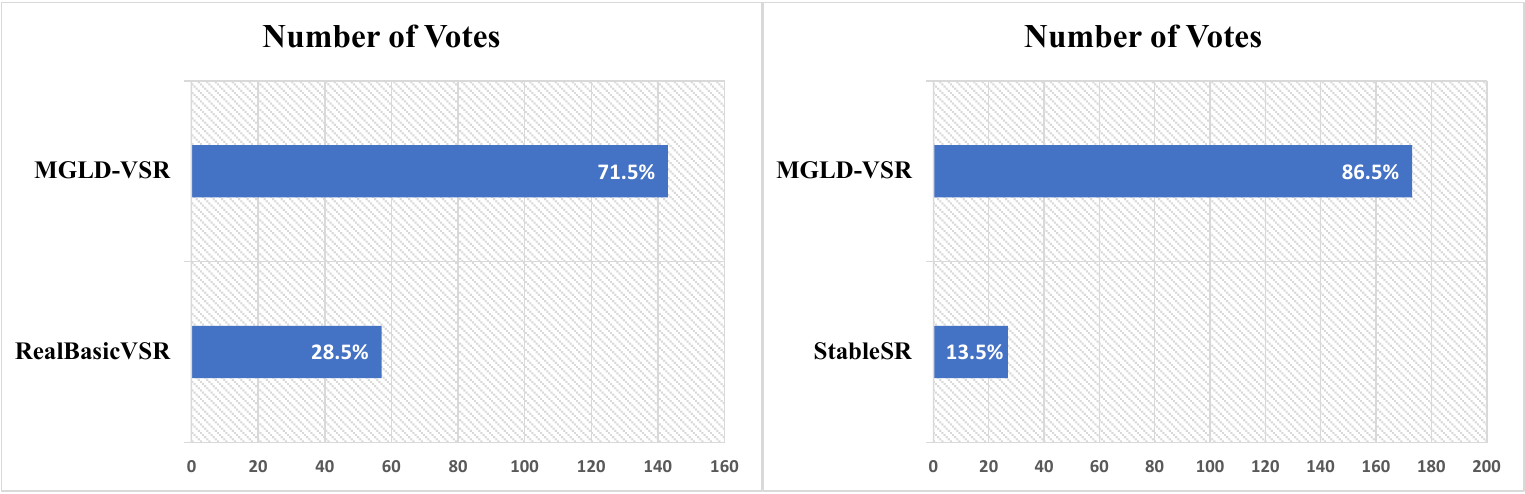}
\caption{User study on 20 videos from VideoLQ evaluated by 10 subjects. MGLD-VSR gains more votes than competing methods.}
\label{figure:user-study}
\end{figure}

\textbf{User Study.}
We also conduct a user study to further examine the effectiveness of the proposed algorithm in comparison with RealBasicVSR \cite{chan2022investigating} and StableSR \cite{wang2023exploiting}. The study is conducted in pairs: given the LR video, the subject is asked to choose the better one from two HR videos obtained by different methods, considering both the reconstruction quality and the temporal consistency. As shown in Figure \ref{figure:user-study}, our method gains more votes in the competitions.

\subsection{Ablation Study}
In this section, we conduct a series of analyses on the proposed framework, including its Motion-guided Diffusion Sampling (MDS) module and Temporal-aware Sequence Decoder (TSD) module, and employ the VideoLQ dataset for experiments. We adopt WE to measure temporal consistency and no-reference metrics NIQE, BRISQUE, MUSIQ and DOVER to measure perceptual quality. 

\begin{table}[!t]
\centering
\caption{Ablation study on the different components of the proposed algorithm. "MDS" refers to Motion-guided Diffusion Sampling and "TSD'' refers to Temporal-aware Sequence Decoder.}
\begin{adjustbox}{width=0.6\columnwidth}
\begin{tabular}{c c | c c c c c}
    \hline
    MDS & TSD  & WE $\downarrow$ & NIQE $\downarrow$ & BRISQUE $\downarrow$ & MUSIQ $\uparrow$ & DOVER $\uparrow$\\ 
    \hline\hline
     &  &  9.3069 & 4.1098 & 25.3249 & 51.5677 & 0.7412 \\
    \checkmark &  & 8.9848 & 3.9876 & 24.0840 & 52.4143 & 0.7435 \\ 
     & \checkmark &  8.3679 & 3.6281 & 22.0806 & 52.6874 & 0.7468 \\ 
    \checkmark	& \checkmark & 7.5478 & 3.5346 & 21.9839 & 52.7812 & 0.7481 \\ 
    \hline
\end{tabular}
\end{adjustbox}
\label{tbl:ablation-component}

\end{table}

\textbf{Different Components of MGLD-VSR.}
As can seen from Table ~\ref{tbl:ablation-component}, the introduction of MDS improves all the metrics over the baseline (i.e., without using MDS and TSD), suggesting that MDS helps not only in reducing the temporal differences (i.e., WE), but also enhancing the perceptual quality of the reconstructed video sequence. 

The incorporation of MDS aids in maintaining perceptual continuity by ensuring that the restored frames could transition smoothly from one to the next. This contributes to more natural and visually pleasing VSR results. The incorporation of TSD also yields better results across all the metrics compared to the baseline. This occurs because TSD is capable of extracting richer contextual information from neighboring frames. When both MDS and TSD are used, the results show further improvements. The full MGLD-VSR approach achieves the best scores on all the metrics, indicating the synergy of the two proposed components for real-world VSR. 

\textbf{Impact of Flow Network.} As shown in Table \ref{tbl:ablation-flownet}, the flow estimation network in motion guided sampling does not show big influence on the results. This might because the motion guided sampling is applied in low-resolution space. 

\begin{figure}[t]
\centering
\includegraphics[width=1.0\linewidth]{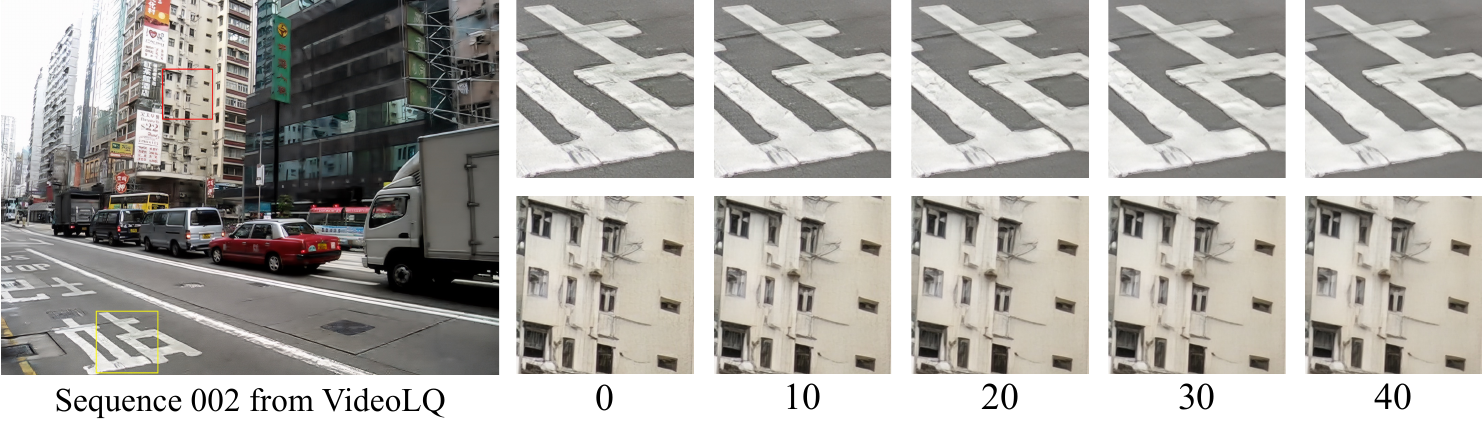}
\caption{Impact of guidance strength during sampling. The numbers below the cropped patches indicate the motion guidance weighting factor $\eta$.}
\label{figure:guidance-scale}
\end{figure}

\textbf{Impact of Guidance Strength.} During guided sampling process, we introduce a weighting factor to control the guidance strength. As shown in Figure \ref{figure:guidance-scale}, a moderate guidance strength could reduce the artifacts and improve the temporal consistency of the generated results. A large guidance strength will blur the results and degrade the reconstructed details. 
To balance the generation power and temporal consistency, we set the weighting factor $\eta = 10$.

\textbf{The Design of VAE Decoder.} To validate the effectiveness of the proposed architecture and loss design for the VAE decoder, we conduct a series of study and the results are shown in Table \ref{tbl:ablation-decoder}. As can be seen from the table, the introduction of the adaptive temporal modeling module improves the temporal consistency as well as perceptual quality (measured by no-reference metrics), and incorporating the feature from encoder to decoder during decoder fine-tuning further improves the results. By introducing two sequence based losses during fine-tuning, we improve the temporal consistency of the results while maintaining or improving the perceptual quality.

\begin{table}[!t]
\centering
\caption{Ablation study on the flow network in motion guided diffusion sampling.}
\begin{adjustbox}{width=0.5\columnwidth}
\begin{tabular}{c | c c c c }
    \hline
    Flow Network & WE $\downarrow$ & NIQE $\downarrow$ & MUSIQ $\uparrow$ & DOVER $\uparrow$ \\ 
    \hline\hline
    SpyNet \cite{ranjan2017optical} & 7.6734 & 3.5545 & 52.6944 & 0.7478 \\
    RAFT \cite{teed2020raft} & 7.5478 & 3.5346 & 52.7812 & 0.7481 \\ 
    \hline
\end{tabular}
\end{adjustbox}
\label{tbl:ablation-flownet}
\end{table}

\begin{table}[!t]
\centering
\caption{Ablation study on the different designs of VAE decoder. "TM" refers to the temporal modeling module. "CFW" refers to the controllable feature warping module that injects VAE encoder feature into VAE decoder.}
    \begin{subtable}[t]{.5\linewidth}
    \begin{adjustbox}{width=1.0\linewidth}
    \begin{tabular}{c | c c | c c c c}
        \hline
        Models & TM & CFW & WE $\downarrow$ & NIQE $\downarrow$ & MUSIQ $\uparrow$ & DOVER $\uparrow$ \\ 
        \hline\hline
        (a) &  &  & 9.8670 & 4.0671 & 50.2707 & 0.7360 \\
        (b) &  & \checkmark & 8.6846 & 3.8451 & 52.5408 & 0.7431 \\ 
        (c) & \checkmark &  & 8.9842 & 3.9236 & 52.4386 & 0.7426 \\ 
        (d) & \checkmark & \checkmark & 7.5478 & 3.5346 & 52.7812 & 0.7481 \\ 
        \hline
    \end{tabular}
    \end{adjustbox}
    \end{subtable}
    \begin{subtable}[t]{.48\linewidth}
    \begin{adjustbox}{width=1.0\linewidth}
    \begin{tabular}{c | c c | c c c c}
        \hline
        Losses & $L_{\mathrm{diff}}$ & $L_{\mathrm{swc}}$ & WE $\downarrow$ & NIQE $\downarrow$ & MUSIQ $\uparrow$ & DOVER $\uparrow$ \\ 
        \hline\hline
        (a) &  &  & 7.9480 & 3.5359 & 52.7894 & 0.7438 \\
        (b) &  & \checkmark & 7.7579 & 3.5409 & 52.5648 & 0.7449 \\ 
        (c) & \checkmark &  & 7.7688 & 3.5376 & 52.7456 & 0.7458  \\ 
        (d) & \checkmark & \checkmark & 7.5478 & 3.5346 & 52.7812 & 0.7481 \\ 
        \hline
    \end{tabular}
    \end{adjustbox}
    \end{subtable}
\label{tbl:ablation-decoder}
\end{table}


\subsection{Model Complexity Analysis}
We provide the comparisons of model parameters and inference time of representative image super-resolution (SR) and VSR methods in Table \ref{tbl:model-detail-1}. Compared with those non-diffusion SR/VSR methods like Real-ESRGAN and RealBasicVSR, MGLD does not have advantages in terms of model parameters and inference time. But compared with diffusion based real-world SR method StableSR, MGLD is much faster in terms of amortized time, benefiting from the lighter CFW design and MIMO design. 
To further improve the speed of MGLD, we can resort to model distillation and more advanced diffusion sampling techniques \cite{salimans2022progressive, meng2023distillation}, which will be our future work.

\begin{table}[htbp]
\centering
\caption{Comparison of model parameters and inference time. The inference time is measured on a NVIDIA A100 GPU for super-resolving a $128 \times 128$ LR input sequence to a $512 \times 512$ HR output sequence. The inference time for RealBasicVSR and MGLD-VSR is amortized from inferring 100 frames. For StableSR and MGLD-VSR, the number of sampling steps is set to 50.}
\begin{tabular}{c | c  c }
    \hline
    Models & Trainable Params. / Total Params.  & Time \\ \hline\hline
    Real-ESRGAN & 16.70M / 16.70M & 0.034s \\
    RealBasicVSR & 6.30M / 6.30M & 0.011s \\
    StableSR & 149.91M / 1.5B & 3.194s \\ 
    MGLD-VSR & 130.45M / 1.5B & 1.113s \\
    \hline
\end{tabular}
\label{tbl:model-detail-1}
\end{table}

\subsection{More Visual Results}  \label{more-visual-results-1}
\begin{figure}[!htbp]
\setlength{\abovecaptionskip}{0.2cm}
\setlength{\belowcaptionskip}{-0.cm}
\centering
\includegraphics[width=1.0\linewidth]{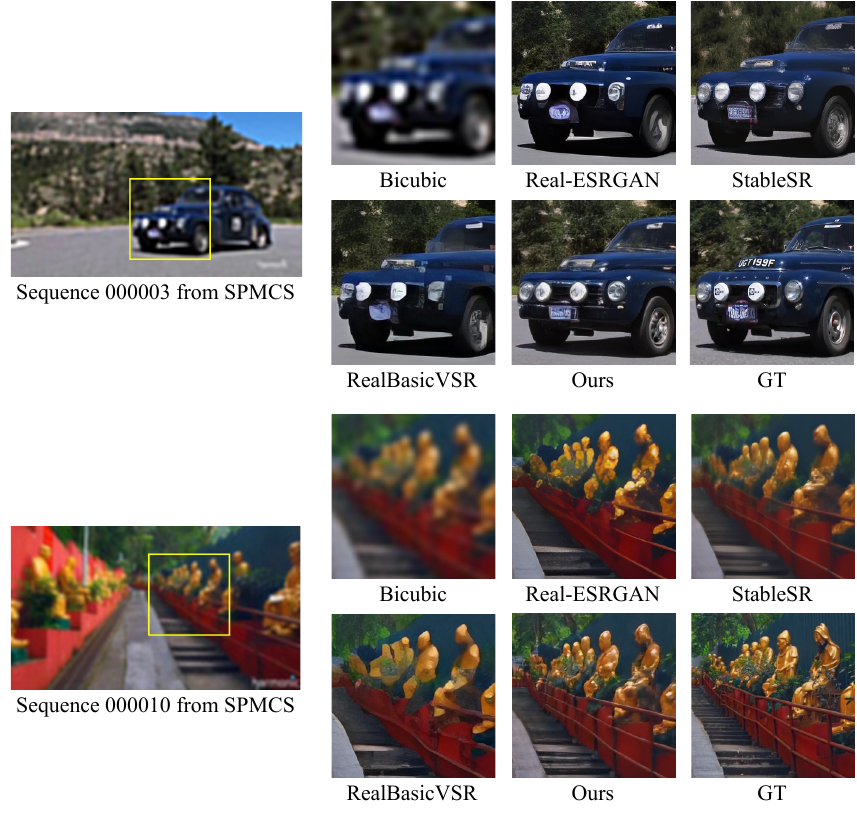}
\caption{Qualitative comparisons on synthetic dataset (SPMCS) for $\times 4$ VSR (Zoom-in for best view).}

\label{figure:visual-comparison-syn-1}
\end{figure}

\begin{figure}[!htbp]
\setlength{\abovecaptionskip}{0.2cm}
\setlength{\belowcaptionskip}{-0.cm}
\centering
\includegraphics[width=1.0\linewidth]{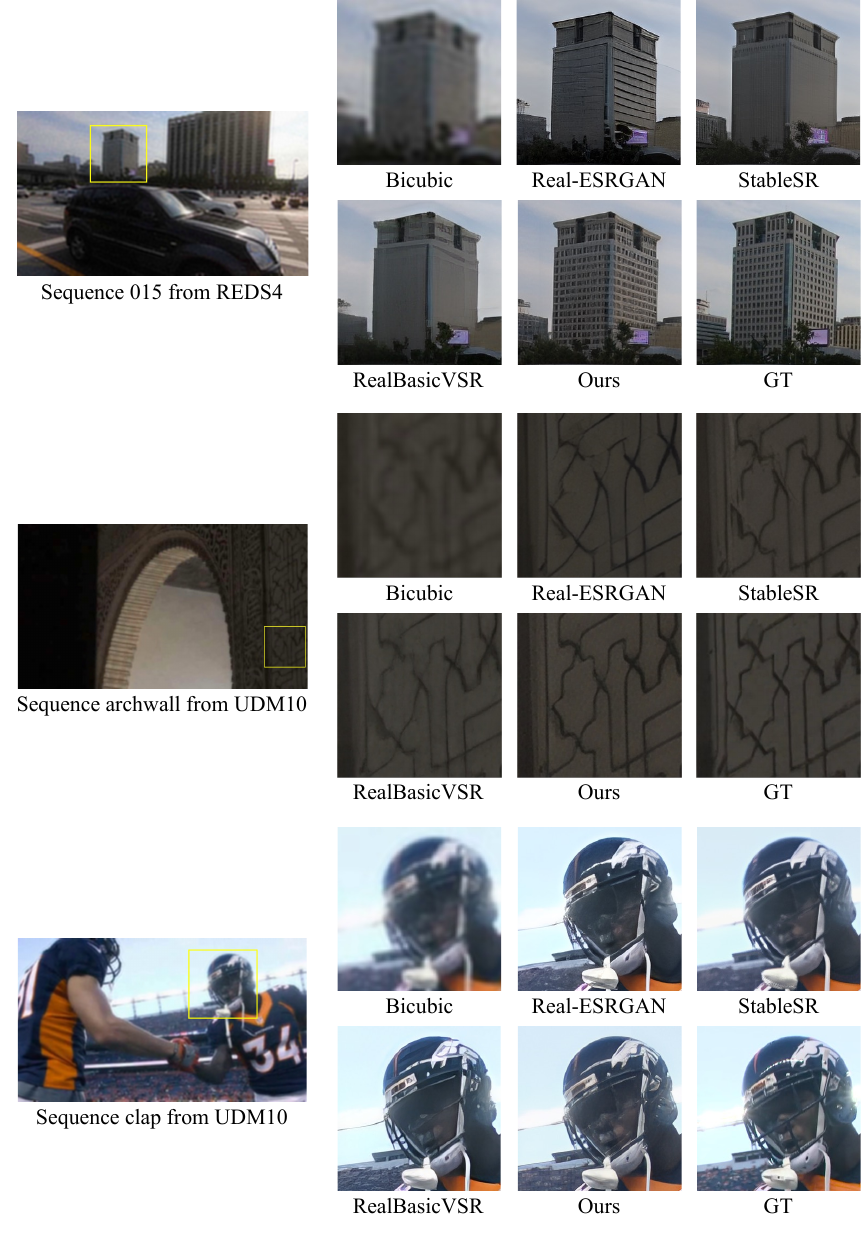}
\caption{Qualitative comparisons on synthetic datasets (REDS4, UDM10) for $\times 4$ VSR (Zoom-in for best view).}

\label{figure:visual-comparison-syn-2}
\end{figure}

In Figure \ref{figure:visual-comparison-syn-1} and Figure  \ref{figure:visual-comparison-syn-2}, we provide more visual results obtained by different algorithms on the synthetic datasets (REDS4, UDM10, SPMCS). One can see that our method can well reconstruct the object structures while generating plausible details under complex degradations.

\begin{figure}[t]
\setlength{\abovecaptionskip}{0.2cm}
\setlength{\belowcaptionskip}{-0.cm}
\centering
\vspace{5ex}
\includegraphics[width=1.0\linewidth]{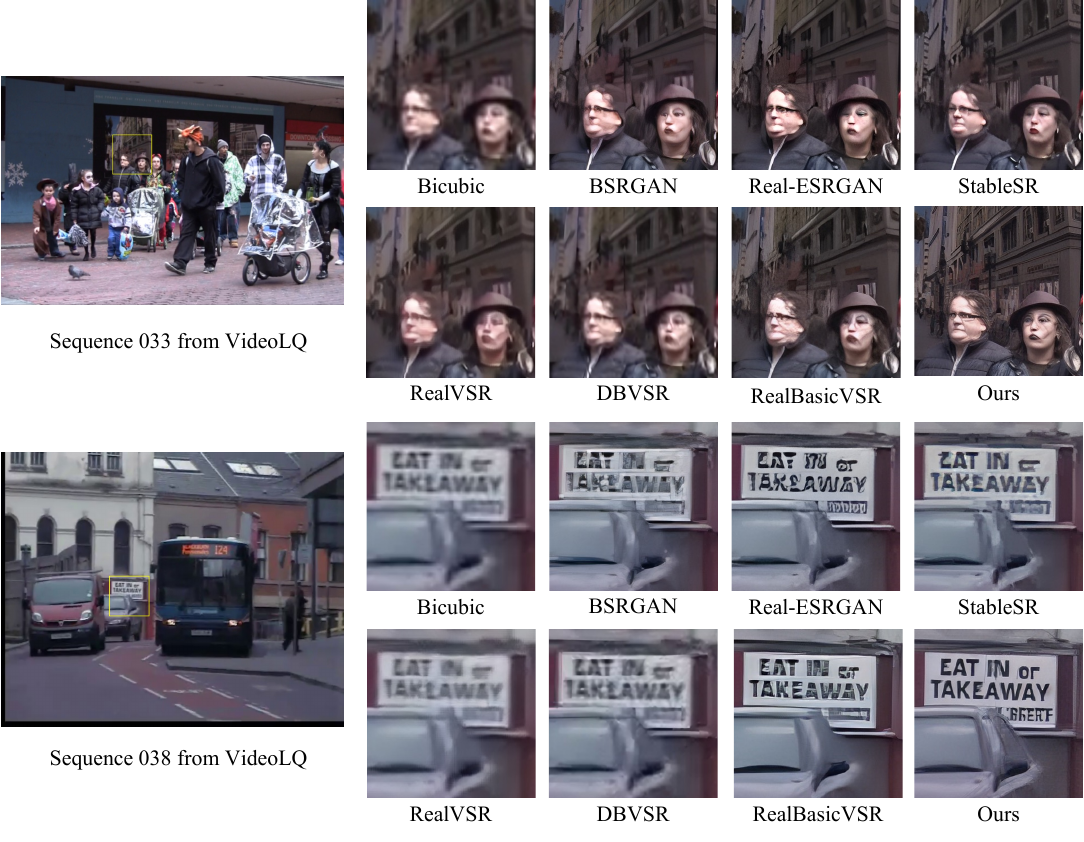}
\caption{Qualitative comparisons on the real-world video dataset (VideoLQ) for $\times 4$ video SR (Zoom-in for best view).}
\label{figure:visual-comparison-real-supp-1}
\end{figure}

\begin{figure}[t]
\setlength{\abovecaptionskip}{0.2cm}
\setlength{\belowcaptionskip}{-0.cm}
\centering
\includegraphics[width=1.0\linewidth]{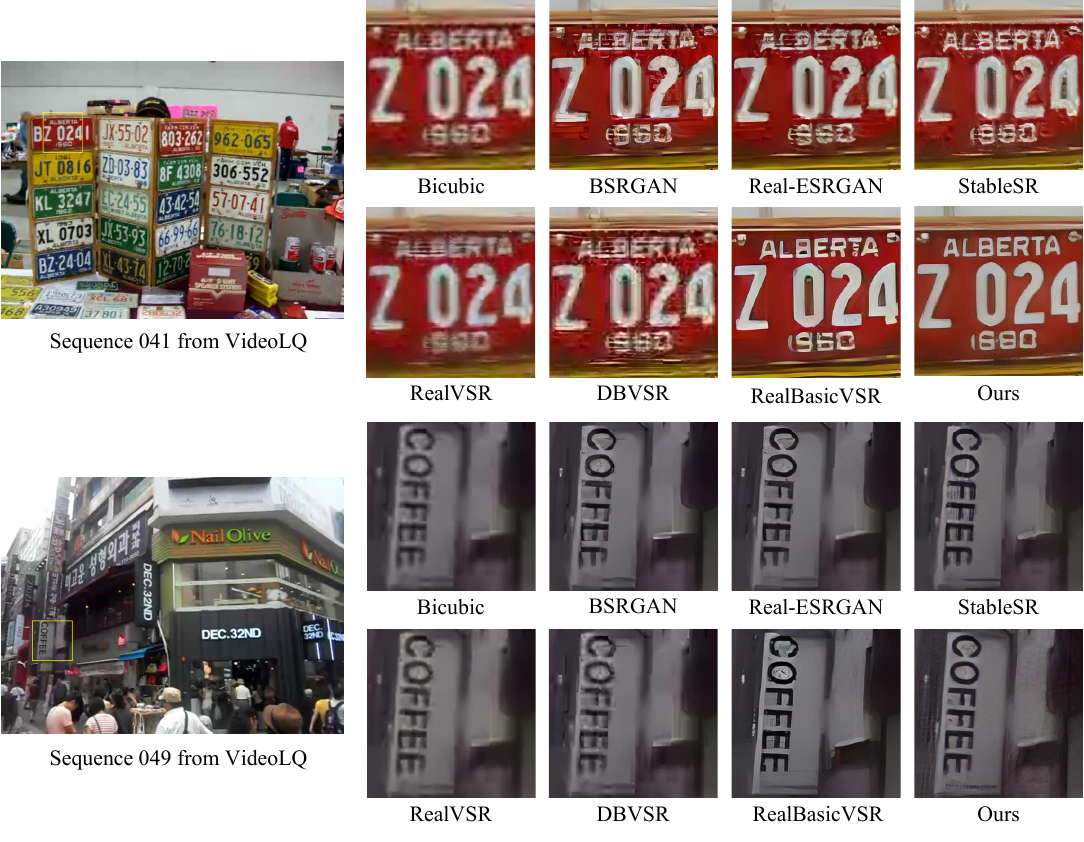}
\caption{Qualitative comparisons on the real-world video dataset (VideoLQ) for $\times 4$ video SR (Zoom-in for best view).}
\label{figure:visual-comparison-real-supp-2}
\end{figure}

In Figure \ref{figure:visual-comparison-real-supp-1} and Figure \ref{figure:visual-comparison-real-supp-2}, we provide more visual results obtained by different algorithms on the real-world video dataset, \ie, VideoLQ. 
As shown by the example of sequence 033 in Figure \ref{figure:visual-comparison-real-supp-1}, MGLD generates clearer plausible structure of the human faces and the background poster. On sequence 038, MGLD clearly reconstructs the characters on the billboard, while most algorithms fail.
As shown in Figure \ref{figure:visual-comparison-real-supp-2}, for sequences 041 and 049 from VideoLQ, MGLD reconstructs clearer characters or numbers while generating less artifacts. 

\subsection{Sequence Demonstrations} \label{more-visual-results-2}

In this section, we provide some sequence demonstrations to validate the advantages of our method. 

\begin{figure}[!htbp]
\centering
\includegraphics[width=1.0\linewidth]{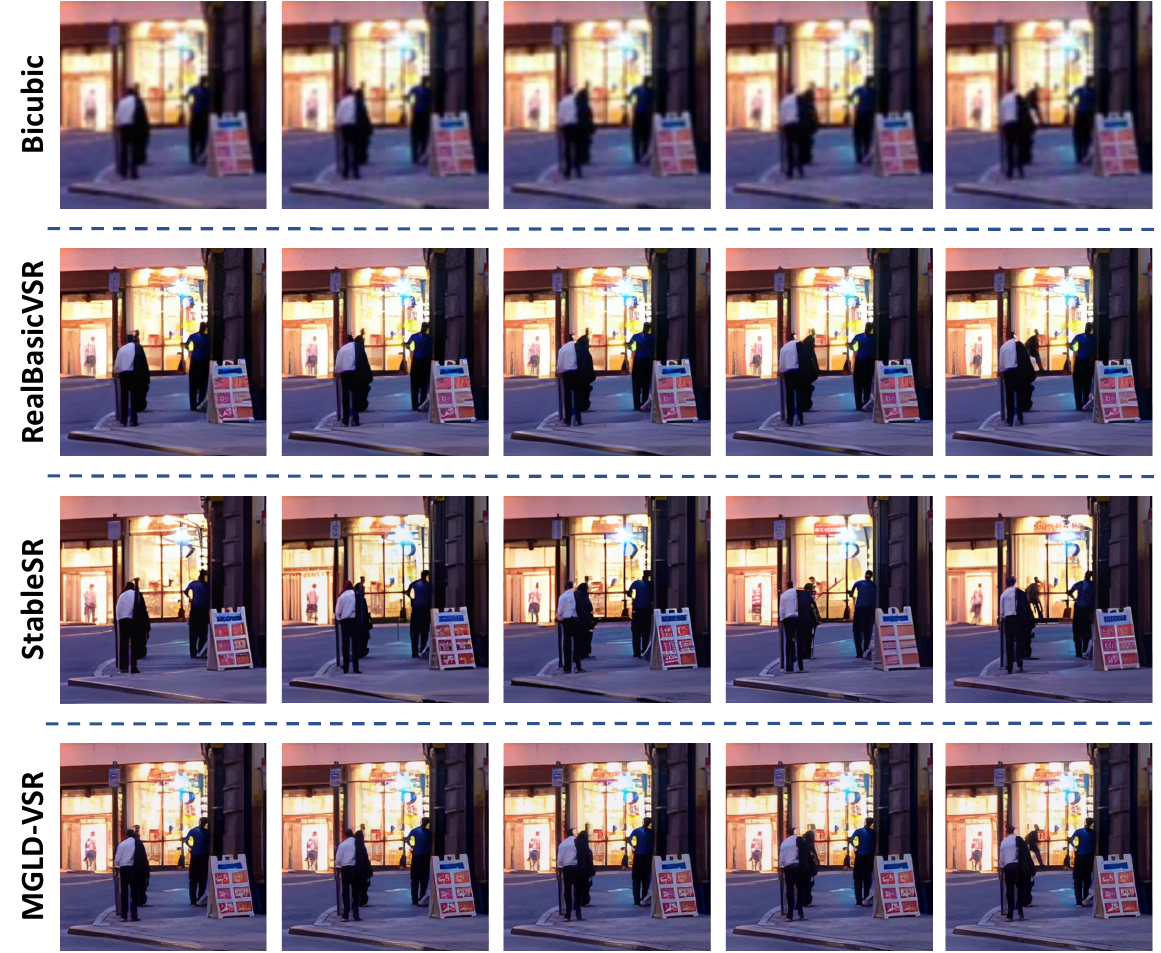}
\caption{Real-world VSR results on sequence 026 from VideoLQ (Zoom-in for best view).}
\label{figure:seq-comparison-real-1}
\end{figure}

\begin{figure}[!htbp]
\centering
\includegraphics[width=1.0\linewidth]{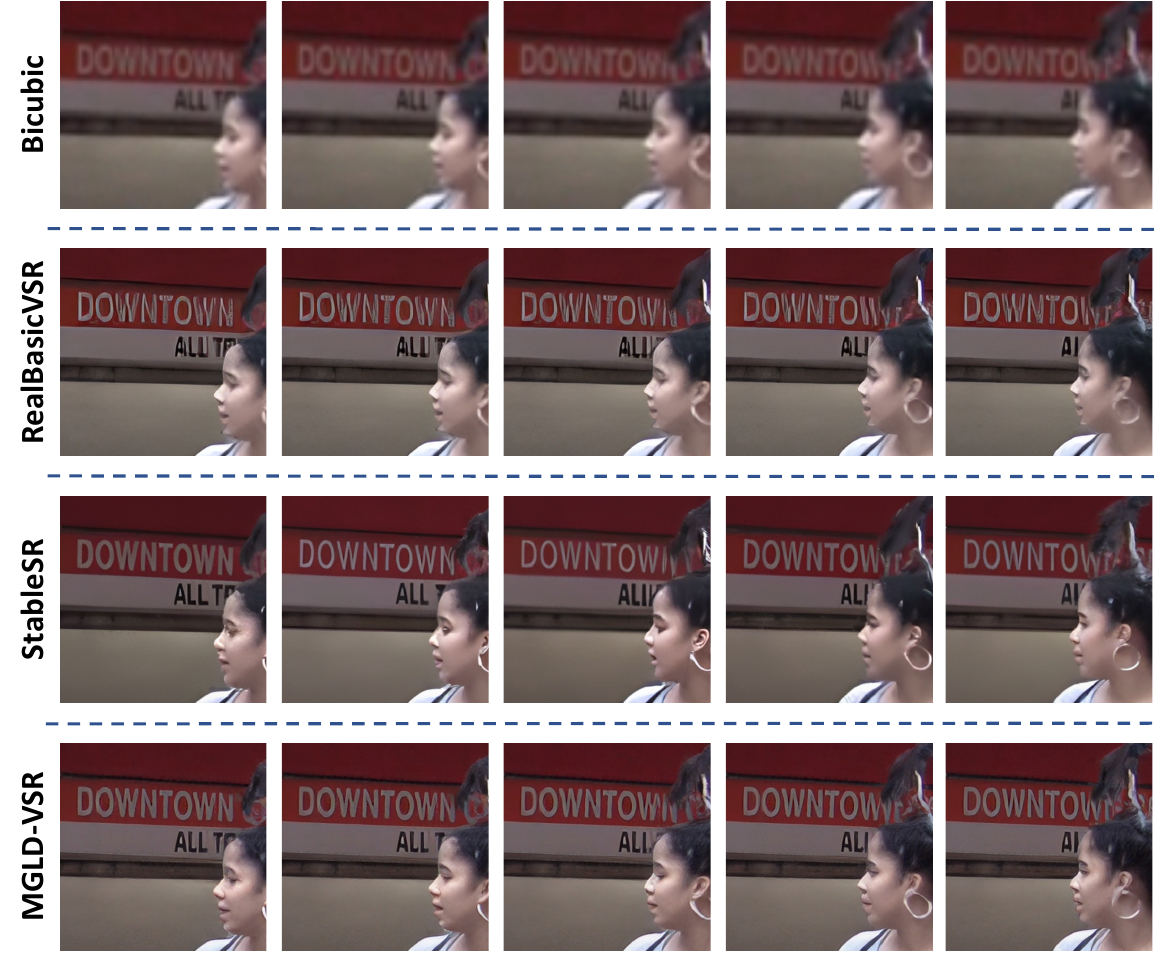}
\caption{Real-world VSR results on sequence 033 from VideoLQ (Zoom-in for best view).}
\label{figure:seq-comparison-real-2}
\end{figure}

\begin{figure}[!htbp]
\centering
\includegraphics[width=1.0\linewidth]{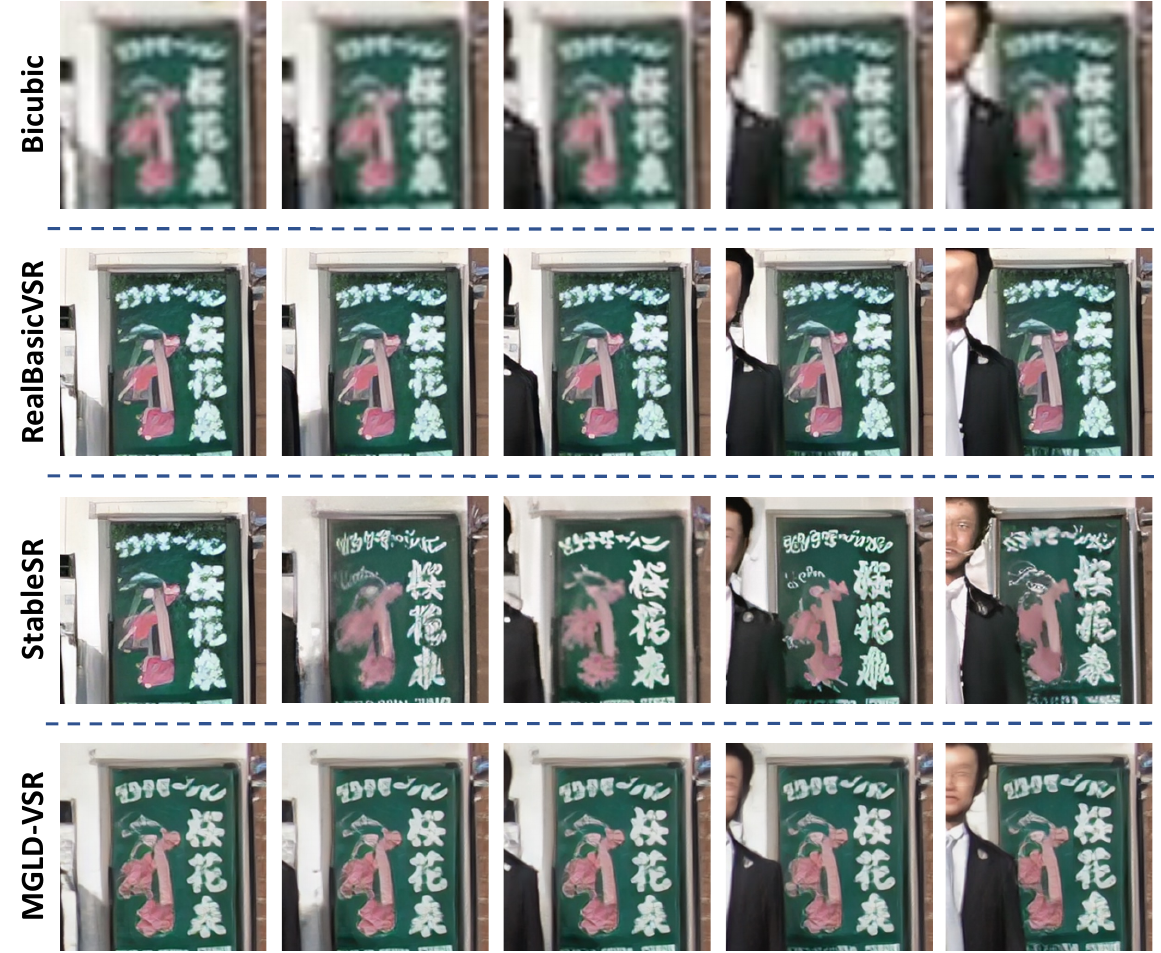}
\caption{Real-world  VSR results on sequence 035 from VideoLQ (Zoom-in for best view).}
\label{figure:seq-comparison-real-3}
\end{figure}

\begin{figure}[!htbp]
\centering
\includegraphics[width=1.0\linewidth]{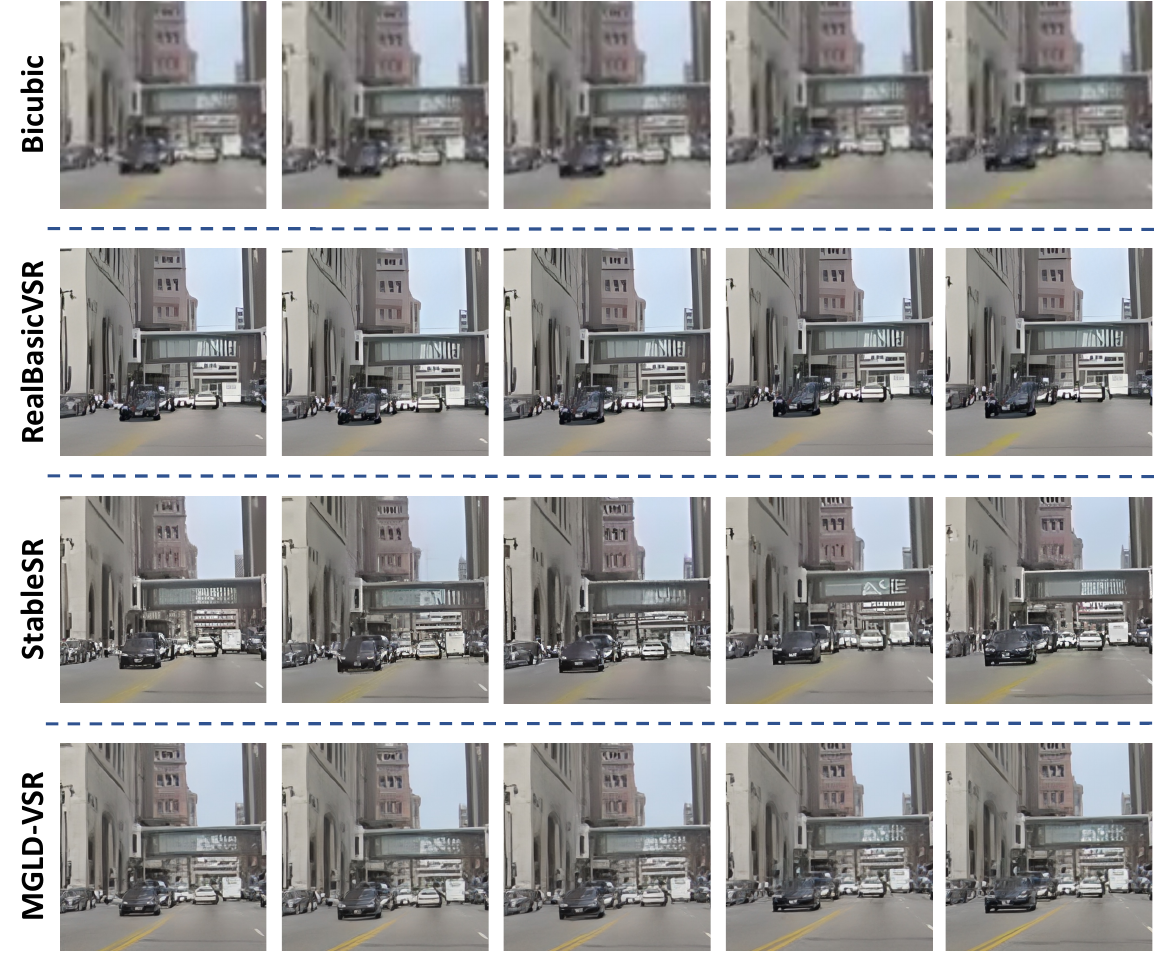}
\caption{Real-world  VSR results on sequence 042 from VideoLQ (Zoom-in for best view).}
\label{figure:seq-comparison-real-4}
\end{figure}

The visual result comparisons on real-world sequences from VideoLQ are shown in Figure \ref{figure:seq-comparison-real-1}, Figure \ref{figure:seq-comparison-real-2}, Figure \ref{figure:seq-comparison-real-3} and Figure \ref{figure:seq-comparison-real-4}. We crop and enlarge the same region from the results obtained by different methods for better comparison. As can be seen from the figures, our method is able to generate realistic details from heavily degraded real-world sequences while achieving good temporal consistency. In comparison, RealBasicVSR fails to generate realistic details and the results obtained from StableSR do not show good temporal consistency.

\section{Conclusion}
We presented a motion-guided latent diffusion algorithm, namely MGLD-VSR, for high-quality and temporally consistent real-world VSR. 
Our method exploited the strong generative prior of pre-trained latent diffusion model to restore video sequence with realistic details. To ensure the content consistency among frames, we developed a motion-guided sampling mechanism, where the motion dynamics of input video were incorporated in the diffusion sampling process. To further improve the continuity of generated details, we introduced a temporal module into the decoder, and fine-tuned it by high quality video sequences with delicately designed sequence-based losses. The fine-tuned sequence decoder mitigated the flickering artifacts and further improved the perceptual quality.
Our proposed MGLD-VSR achieved state-of-the-art visual results on the widely adopted real-world VSR benchmarks, validating the great potentials of latent diffusion priors in enhancing real-world video quality.

\clearpage  


%
%
\bibliographystyle{splncs04}
\bibliography{main}
\end{document}